\pdfoutput=1

\documentclass[11pt]{article}

\usepackage[]{acl}

\usepackage{times}
\usepackage{latexsym}

\usepackage[T1]{fontenc}

\usepackage[utf8]{inputenc}

\usepackage{microtype}

\usepackage{subcaption}
\usepackage{graphicx}
\usepackage{booktabs}
\usepackage{multirow}
\usepackage{amsmath}
\usepackage{algpseudocode}
\usepackage{amsfonts}
\usepackage{amssymb,graphicx}
\usepackage{makecell}
\usepackage{algorithm}
\usepackage{tabularx}
\usepackage{pifont}
\newcommand{\cmark}{\ding{51}}%
\newcommand{\xmark}{\ding{55}}%
\usepackage{csquotes}

\usepackage{soul,xcolor}
\setstcolor{red}

\title{Efficient Multi-Agent Collaboration with Tool Use \\for Online Planning in Complex Table Question Answering}

\usepackage{todonotes}
\definecolor{mohsen}{rgb}{0.635,0.998,0.722}
\definecolor{anne}{rgb}{0.8,0.8,1}
\definecolor{wei}{rgb}{0.998,0.722,0.635}
\definecolor{heike}{rgb}{0.4, 0.8, 0.4}
\definecolor{darkgreen}{rgb}{0, 0.36, 0.608}
\definecolor{final}{rgb}{1, 1, 0.6}

\newcommand{\anne}[2][black]{\textcolor{#1}{#2}}

\author{Wei Zhou$^{1,2}$ \hspace{5mm}
  Mohsen Mesgar$^1$ \hspace{5mm}
  Annemarie Friedrich$^2$\hspace{5mm} 
   Heike Adel$^{3}$ \\
  $^1$Bosch Center for Artificial Intelligence, Renningen, Germany \\ 
     $^2$University of Augsburg, Germany \hspace{2.0mm} $^3$Hochschule der Medien, Stuttgart, Germany \hspace{5mm} \\
\texttt{\{wei.zhou|mohsen.mesgar\}@de.bosch.com}\\ 
  \texttt{annemarie.friedrich@uni-a.de} \hspace{5mm} \texttt{adel-vu@hdm-stuttgart.de}}

\begin{document}
\maketitle

\begin{abstract}
Complex table question answering (TQA) aims to answer questions that require complex reasoning, such as multi-step or multi-category reasoning, over data represented in tabular form.
Previous approaches demonstrate notable performance by leveraging either closed-source large language models (LLMs) or fine-tuned open-weight LLMs. 
However, fine-tuning LLMs requires high-quality training data, which is costly to obtain. The use of closed-source LLMs poses accessibility challenges and leads to reproducibility issues. 
In this paper, we propose \textbf{M}ulti-\textbf{A}gent \textbf{C}ollaboration with \textbf{T}ool use (MACT), a framework that requires neither fine-tuning nor closed-source models. 
In MACT, a planning agent and a coding agent that also make use of tools collaborate for TQA. 
MACT outperforms previous SoTA systems on three out of four benchmarks and performs comparably to the larger and more expensive closed-source model GPT-4 on two benchmarks, even when using only open-weight models without any fine-tuning. 
Our extensive analyses prove the effectiveness of MACT's multi-agent collaboration in TQA.
We release our code publicly.\footnote{\url{https://github.com/boschresearch/MACT}}


\end{abstract}

  \section{Introduction}
\label{sec:intro}
\begin{figure}[!t]
    \centering \includegraphics[width=1.0\columnwidth]{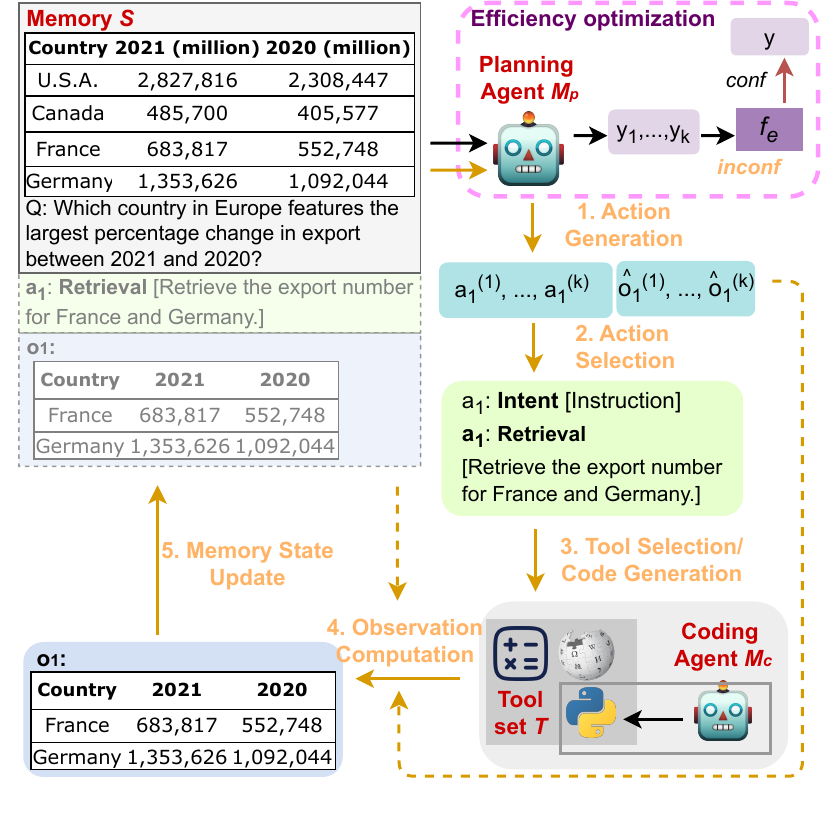}
    \caption{
    Overview of MACT, an iterative collaboration framework for TQA that consists of five stages for each iteration as well as an efficiency optimization module.} 
    \label{fig:method}
\end{figure}

The goal of table question answering (TQA) is to answer a question based on data represented in tabular form, optionally also using additional textual context.
Recent studies on TQA focus more and more on complex instances, as they are ubiquitous in table data analysis \citep{zhu-etal-2021-tat,Zhang2024TableLLMET,lu-etal-2023-scitab}.
Solving those complex instances requires performing multiple reasoning steps and/or employing different reasoning strategies \citep{ghosal-etal-2023-retag}. We refer to these aspects as \textit{multi-step} and \textit{multi-category} reasoning, respectively.
An example requiring both types of reasoning is shown in the upper left part of Figure \ref{fig:method}.
To answer the question about the percentage change, a system first needs to use factual knowledge to extract countries in Europe. 
Then, numerical reasoning is applied to calculate the percentage change and to carry out the comparison.

One popular approach for addressing those complex instances in TQA is planning, where step-wise plans are generated and used to guide the reasoning process \citep{zhang2023reactable, wang2024chain, Wu2024ProTrixBM, Zhu2024TATLLMAS, zhao-etal-2024-tapera}.
State-of-the-art works in this direction either fine-tune open-weight large language models (LLMs) \cite{Wu2024ProTrixBM, Zhu2024TATLLMAS} or prompt closed-source commercial LLMs \cite{wang2024chain, zhang2023reactable, zhao-etal-2024-tapera}.
However, \textbf{fine-tuning requires high-quality data}, which is usually expensive to obtain \citep{zhu-etal-2021-tat}.
Prompting closed-source commercial LLMs can also be \textbf{costly and poses challenges to reproducibility}.
To the best of our knowledge, existing methods leverage a single LLM to perform planning and reasoning, which is sub-optimal in particular if the LLM does not excel at mathematical reasoning or coding \citep{Wu2024ProTrixBM}. 
These models struggle with answering questions requiring complex reasoning.

To address these challenges, we propose MACT, a multi-agent collaboration framework with tool use, which neither depends on closed-source LLMs nor requires fine-tuning.
In fact, its backbone LLMs can be exchanged flexibly. 
It incorporates two agents (a planning agent and a coding agent) and a set of tools (a Python interpreter, a calculator and Wikipedia search). 
The planning agent performs online planning, i.e., it generates a plan iteratively.
This breaks down complex problems and helps to address multi-step reasoning.
The coding agent and the tool set assist with generating faithful intermediate results.
The agents work in a collaborative setting, addressing the challenges of multi-category reasoning as all agents can concentrate on the reasoning types they excel in.
An efficiency optimization module which allows the framework to take informed shortcuts.

We conduct experiments on four popular TQA benchmarks that include complex TQA instances.
Our framework outperforms previous SoTA systems on three out of four benchmarks.
It achieves comparable results to GPT-4 on two benchmarks even when using only open-weight models without any fine-tuning.
In comparison to fine-tuned SoTA TQA systems, it demonstrates considerably better generalizability across datasets.
Our analysis proves the effectiveness of our proposed collaborative setting of specialized agents. 
We find that the efficiency optimization module can save up to 33\% of iterations without performance degradation.

 \section{Related Work}
\label{sec:intro}
We review previous work for three core aspects of MACT: planning, multi-agent collaboration and LLMs with tool use.
Table \ref{tab:compare_framework} compares MACT with previous TQA systems.

\begin{table}[!htb]
    \small
    \centering
    \begin{tabularx}{.47\textwidth}{l p{0.6cm} p{0.5cm} p{0.4cm} p{0.4cm} X}
         \toprule
         System & \rotatebox{45}{\makecell{Online\\ planning}} & \rotatebox{45}{\makecell{\ No fine-  \\ tuning}}& \rotatebox{45}{NL plan}& \rotatebox{45}{\makecell{Multi \\ agents}}& Tools\\\midrule
         Raven &\textcolor{red}{\xmark} & - & - & \textcolor{red}{\xmark} & cal, SQL \\
         Binder &\textcolor{red}{\xmark}  & - & - & \textcolor{red}{\xmark}& SQL/Python   \\
         Lever &\textcolor{red}{\xmark}  & - &- & \textcolor{red}{\xmark} &SQL \\
         TableLlaMA & \textcolor{red}{\xmark}  & - &- & \textcolor{red}{\xmark}&\textcolor{red}{\xmark}\\
         Dater & \textcolor{red}{\xmark} & \textcolor{green}{\cmark} & \textcolor{green}{\cmark}&\textcolor{red}{\xmark} &\textcolor{red}{\xmark} \\
         TAT-LLM & \textcolor{red}{\xmark} &\textcolor{red}{\xmark} &\textcolor{green}{\cmark} &\textcolor{red}{\xmark}& cal  \\
         Chain of table &\textcolor{green}{\cmark}  & \textcolor{green}{\cmark}  & \textcolor{red}{\xmark}& \textcolor{red}{\xmark} & \textcolor{red}{\xmark}\\
         Reactable & \textcolor{green}{\cmark} & \textcolor{green}{\cmark}& \textcolor{red}{\xmark} &\textcolor{red}{\xmark} &SQL, Python\\
        Protrix & \textcolor{green}{\cmark}& \textcolor{red}{\xmark} &  \textcolor{green}{\cmark} &\textcolor{red}{\xmark}& SQL \\
        TAPERA & \textcolor{green}{\cmark} &  \textcolor{green}{\cmark} & \textcolor{green}{\cmark} & \textcolor{red}{\xmark} & Python \\
         MACT (ours) &\textcolor{green}{\cmark} &  \textcolor{green}{\cmark}&\textcolor{green}{\cmark} & \textcolor{green}{\cmark} & Python, cal, Wiki \\
    \bottomrule
         
    \end{tabularx}
    \caption{Comparing MACT with previous works. NL plan stands for using natural language to encode a plan, cal=calculator, and Wiki=Wikipedia.}
    \label{tab:compare_framework}
\end{table}

\paragraph{Planning.} 
We categorize previous work into three groups based on planning strategies: \textit{Heuristic coarse-grained planning} consists of  two pre-defined steps of retrieving and aggregating \cite{Ye2023LargeLM, Zhou2024FREBTQAAF}.
 \textit{Online global planning} generates a plan in the first iteration and revises it in the next one \cite{zhao-etal-2024-tapera}. \textit{Online iterative planning} conditions the generation of the next step on the executed results of previous steps \cite{Zhang2023ReAcTableER,wang2024chain}.
We opt for online iterative planning for complex TQA, as it 
provides for more fine-grained steps during problem solving 
(in contrast to heuristic planning) and emphasizes the dependency among steps (in contrast to online global planning), which is crucial in complex TQA. 
In contrast to previous work using iterative planning, we introduce an efficiency optimization module to minimize the costs of the framework.
To learn how to generate plans, previous work either depends on fine-tuning \citep{Wu2024ProTrixBM, Zhu2024TATLLMAS}, or strong closed-source models, combined with in-context learning \citep{wang2024chain, Zhang2023ReAcTableER, zhao-etal-2024-tapera}. 
By contrast, MACT generates effective plans using either closed-source or open-weight models, without the need for fine-tuning. 

\paragraph{Multi-agent collaboration.} In multi-agent collaboration settings, multiple AI entities collaborate towards a common goal \citep{talebirad2023multiagentcollaborationharnessingpower}.
We use the term agent to refer to LLMs that interact with executable tools, following \citet{Qiao2024AutoActAA}.
As far as we know, none of the previous works in TQA utilize multi-agent collaboration.
The approaches most closely related to ours explore collaboration among 
homogenous agents, i.e., all agents use the same backbone but are prompted differently \citep{Liu2023RethinkingTD, zhao-etal-2024-tapera}.
The effectiveness of this approach relies heavily on strong (closed-source) backbone models. 
Our work explores multi-agent collaboration for TQA, without any constraint of model type.

\paragraph{Tool use.}
LLMs have been shown to be ineffective in retrieving information from long tables \citep{Zhou2024FREBTQAAF} and carrying out numerical reasoning \citep{imani2023mathpromptermathematicalreasoningusing}.
Making use of tools can ensure faithful results of these operations.
The most common tools used in TQA are SQL interpreters \cite{Cheng2022BindingLM}, Python with Pandas dataframes \cite{gemmell-dalton-2023-toolwriter}, and calculators \citep{Zhu2024TATLLMAS}. 
In MACT, we use similar tools.
Inspired by \citet{Shinn2023ReflexionLA}, we further add Wikipedia search as an additional tool to assist questions requiring factual knowledge.

\section{Method}
\label{sec:method}

We propose MACT, a \textbf{M}ulti-\textbf{A}gent \textbf{C}ollaboration framework enriched with a set of \textbf{T}ools for TQA.
Figure \ref{fig:method} provides an overview of the framework.
It consists of four major modules:
a memory $S$,
a planning agent $M_p$,
a coding agent $M_c$, and
a tool set $T$.
$M_p$ and $M_c$ are instantiated by (potentially) different LLMs.
They collaborate through five core stages:
\textit{action generation},
\textit{action selection},
\textit{tool selection/code creation},
\textit{observation computation}, and
\textit{memory state update}. 
The stages are executed iteratively for a maximum of $I$ iterations, where $I$ is a hyper-parameter.
We control the overall efficiency of the collaboration via an \emph{efficiency optimization} module. 
For a new TQA instance, we initialize the memory state $s_0 = (\text{table}, \text{question}, \text{texts})$, i.e., with the input table, the question, and (if given) textual context.
All parts of the memory are represented as strings. To represent the table as a string, we use pipes as column separators.

\subsection{Action Generation}
To format our plans, we follow ReAct \citep{yao2023react} that consists of the generation of thoughts, actions and observations. 
Our framework only requires actions and observations, but following \citet{yao2023react}, who demonstrate performance gains from generating thoughts with actions, we adopt their prompting method.
Thus, at each iteration $i \le I$, we prompt $M_p$ to generate a thought $z_i$, an action $a_i$ and an observation $\hat{o}_i$: 
$(z_i, a_i, \hat{o}_i) \sim M_p(z_i, a_i, \hat{o}_i|s_{i-1}, \phi_p, \tau_p)$, where $s_{i-1}$ is the previous memory state, and $\phi_p$ and $\tau_p$ are the prompt (provided in \ref{prompts}) and temperature of the LLM used for $M_p$, respectively.
Note that $\hat{o}_i$ is not the final observation.
Instead, it is later used during execution as a particular form of our proposed collaboration between the planning and coding agent (see \ref{execution}). 
We sample from $M_p$ $k$ times, resulting in $k$ actions $\{a_{i}^n\}_{n\le k} =\{a_{i}^{1}, a_{i}^2, ..., a_{i}^k\}$ and their corresponding estimated observations $\{\hat{o}_{i}^n\}_{n\le k} = \{\hat{o}_{i}^1, \hat{o}_{i}^2, ..., \hat{o}_{i}^k\}$ in iteration $i$. 
Following \citet{yao2023react}, we define an action $a_i$  with two parts: an intent and an instruction, e.g., \enquote{\textit{Retrieval} [Retrieve the export number for France and Germany].}
The intent encodes the purpose of an action, e.g., \textit{Retrieval} denotes retrieving information from the input table. The instruction (marked with brackets) provides detailed specifications of the intent.
Table \ref{tab:actions} shows the six types of intents we define for our framework and examples for corresponding instructions.

\begin{table*}
\footnotesize
\begin{tabularx}{\textwidth}{@{}lXp{2.5cm}lp{4.5cm}@{}}
\toprule
    \textbf{Intent} & \textbf{Instruction: Format and Example} & \textbf{Tool Selection} & \textbf{Code Generation} & \textbf{Tool Use/Execution} \\
    \midrule
    \textit{Retrieval} & textual description of what to retrieve, e.g., ``sale numbers of 2019'' & $t=Python$ & $M_c$& Python interpreter is run on generated code\\
    \textit{Calculation} & formula or textual description of what to calculate, e.g., ``(135-114)/135''   & $t=T_{cal}$ if formula, else $t=T_{python}$ & $M_c$ if not formula & Calculator is executed or Python interpreter is run on code\\
    \textit{Search} & entity name, e.g., ``Tesla'' & $t=T_{srh}$ & no&Wikipedia API is called on entity\\
    \textit{Read} & textual description of required information from input texts, e.g., ``when was the target method adopted'' & $t=Null$ &no &$M_p$ is prompted to extract information from provided textual context\\
     \textit{Finish} & final answer $y$ & $t=Null$ & no&final answer is output and execution stops\\
   \textit{Ask} & textual description of required information from $M_p$, e.g., ``hours of a day'' & $t=Null$ & no&$M_p$ is prompted for the information need\\
    \bottomrule
\end{tabularx}
\caption{Overview of intents and instructions of actions and how they are executed within our framework.}
\label{tab:actions}
\end{table*}

The intents \textit{Retrieval} and \textit{Calculation} are commonly seen in previous works \citep{gemmell-dalton-2023-toolwriter, Zhu2024TATLLMAS}. 
We use \textit{Retrieval} for any operations extracting information from a table, including  direct querying, filtering, and grouping.
Instructions that require calculation, counting or comparison are captured by \textit{Calculation}.
To fulfill the possible need for external (factual) knowledge that is not present in the table or textual context, we add the intent \textit{Search}, which performs Wikipedia searches to retrieve informative text passages.
\textit{Read} covers the need for contextual reasoning in table-text QA. 
It refers to instructions involving retrieving information from the texts provided as part of TQA instances. 
The intent \textit{Finish} stops $M_p$ from generating more actions and ends the iterative execution of our framework, providing the final answer in the corresponding instruction. 
Lastly, we use an intent called \textit{Ask} to retrieve an answer based on the internal knowledge of the planning agent. 
If $M_p$ fails to generate a valid action at an iteration, it will continue to generate the action at the next iteration until reaching the maximum iteration number $I$, and return the most common prediction directly from $M_p$ as the final answer (See section \ref{sec:efficiency}). 

\subsection{Action Selection}
From the set of $k$ actions generated by $M_p$ for iteration $i$, we use a function $f_s$ to select the most promising action $a^*_i=f_s(s_{i-1}, \{a_{i}^n\}_{n\le k})$.
We use self-consistency (\texttt{SC}) \citep{wang2023selfconsistencyimproveschainthought} 
as the selection function, which outputs the most frequent action from the set of sampled actions. 
In the case of ties, we choose the most frequent action that was sampled first.
We provide a comparison with other selection functions in \ref{selection_model}. 


\subsection{Tool Selection and Use} 
The tool needed for executing action $a^*_i$ depends on the intent of the action (see columns \enquote{Tool Selection} and \enquote{Tool Use/Execution} of Table \ref{tab:actions}).
 To address the intents \textit{Search}, \textit{Calculation} and \textit{Retrieval}, we introduce a set of tools $T = \{T_{srh}, T_{cal}\, T_{python}\}$.
 $T_{srh}$ is an API function for Wikipedia search \citep{Shinn2023ReflexionLA} from Langchain.\footnote{\url{https://rubydoc.info/gems/langchainrb/Langchain/Tool/Wikipedia}}
 The API takes a target entity specified in the instruction and returns the first paragraph of the corresponding Wikipedia entry. 
 $T_{cal}$ is a calculator, powered by a Python interpreter. 
 It takes a formula generated by $M_p$ and outputs the answer.
 Note that the instruction of \textit{Calculation} can also be a textual description, such as \enquote{Compute the average number of medals for each country in the table.}
 To better address these instructions, we introduce a coding agent $M_c$ with a Python interpreter, denoted as $T_{python}$ (see \ref{code_gen_exe}). 
 We do not distinguish formulas and text description. This means any instructions with the intent \textit{Calculation} will be firstly passed to $T_{cal}$. If $T_{cal}$ fails to execute the instruction, $T_{python}$ is applied.
 
The intent \textit{Retrieval} is also addressed by $M_c$ and $T_{python}$, i.e., $M_c$ generates Python code based on a given instruction to retrieve target cells in the tables and the Python interpreter returns the executed results.
 Lastly,  For \textit{Read}, \textit{Ask} and \textit{Finish}, no tool is used, denoted as $t=Null$ in Table \ref{tab:actions}.
 The answers to the intents \textit{Read} and \textit{Ask} are queried via $M_p$. For \textit{Read}, $M_p$ reads the answer from a given textual context. For \textit{Ask}, it responds based on its internal knowledge. 
 No execution is performed for \textit{Finish}, as the intent ends MACT with a final answer in its instruction.

\subsection{Code Generation and Execution}
\label{code_gen_exe}
To address textual instructions for \textit{Calculation} actions as well as \textit{Retrieval} actions, we integrate a coding agent $M_c$, which is an LLM and translates the instructions of $a^*_i$ into Python code snippets $c_i \sim M_c(c_i|a^*_i,s_{i-1}, \phi_c, \tau_c)$.
 The hyper-parameter $\tau_c$ controls the temperature of the coding agent, and $\phi_c$ is a static, pre-defined prompt (see \ref{prompts}). 
We sample $k$ times from $M_c$ to increase the robustness of the system against generated syntax errors, resulting in a set of code snippets $C = \{c_{i}^n\}_{n\le k}$.
A Python interpreter is run on each $c_i$, creating a set of executed solutions $\hat{C} = \{\hat{c}_{i}^n\}_{n\le k}$ .

\subsection{Observation Computation}
\label{execution}
The computation of the final observation $o_i$ depends on the selected tool $t_i$: if $t_i \in \{T_{cal}, T_{srh}\}$, the corresponding tool returns a deterministic result.  
If $t_i$ is $T_{python}$, 
we select the most frequent element from the combined set $\{\hat{o}_{i}^n\}_{n\le k} \cup \hat{C}$ as final observation, where $\{\hat{o}_{i}^n\}_{n\le k}$ is the estimated observations sampled from $M_p$.
This strategy features two levels of collaboration: from an ensemble perspective, both $M_p$ and $M_c$ contribute to obtain $o_i$; from a pipeline perspective, $M_c$ makes use of the outputs (actions) of $M_p$.
If neither $T$ nor $M_c$ are needed to execute the action, the final observation is the most frequent element in $\{\hat{o}_{i}^n\}_{n\le k}$.



\subsection{Memory State Update and Iteration}
After obtaining $o_i$, we update the memory state with the selected action and the observation of iteration $i$: $s_i = s_{i-1} + [a^*_i, o_i]$.
The framework then continues with the next iteration $i+1$.
Note that adding the observation to the memory state also allows $M_p$ to build on top of results from $M_c$ in iteration $i+1$.
If $i>I$, the execution stops with a predicted answer directly from $M_p$. In 98\% of the cases, a final answer is given before $i>I$.

\subsection{Efficiency Optimization}
\label{sec:efficiency}
The iterative collaboration approach of $M_p$ and $M_c$ is highly effective in practice as demonstrated in our experiments.
However, questions that do not require multi-step or multi-category reasoning can also be answered directly by $M_p$.
For those instances, we propose an efficiency optimization component that serves as a shortcut for directly outputting an answer in the first iteration.
Whether the answer is output directly depends on the confidence of $M_p$, which we approximate by the degree of self-consistency of its estimated predictions $Y = \{y^1, ..., y^k\}$. 
$Y$ is obtained by accessing the whole reasoning trace (consisting of $j$ actions and estimated observations) that $M_p$ generates for a given $s_0$ until the intent \textit{Finish} is output: i.e., $(a^k_1, \hat{o}^k_1, ... a^k_j, \hat{o}^k_j) \sim M_p(a^k_1, \hat{o}^k_1, ... a^k_j, \hat{o}^k_j |s_0, \phi_p, \tau_p)$. 
The output $y^k$ is the instruction of the action $a^k_j$ with intent \textit{Finish}. 
To control the trade-off between performance and computation time, we introduce a hyper-parameter $\alpha \in [0..1]$. 
If the degree of self-consistency, i.e., the number of occurrences of the most frequent prediction in $Y$ is larger than $\alpha*k$ (high degree of \texttt{SC}), $M_p$ outputs this most frequent answer. Otherwise, the collaborative framework as described before is adopted.
In short, the smaller $\alpha$, the more often the system is allowed to use the shortcut, and the larger $\alpha$, the more confident $M_p$ needs to be in order to take the shortcut. 
 \section{Experiments}
\label{sec:exp}
We assess the performance of MACT on four TQA benchmarks in comparison to SoTA TQA systems.

\paragraph{Datasets.}
We choose four TQA datasets that cover different reasoning complexities and domains (See \ref{datasets} for more details).
\textbf{WTQ} \cite{Pasupat2015CompositionalSP} is the easiest dataset as it neither requires multi-step nor multi-category reasoning. 
However, it is a widely used benchmark for TQA in the general domain and enables a fair comparison with recent TQA systems. 
\textbf{TAT} \citep{zhu-etal-2021-tat} includes hybrid tabular and textual data.
Most questions require numerical reasoning.
\textbf{CRT} \citep{zhang-etal-2023-crt} uses Wikipedia tables and involves complex reasoning. 
\textbf{SCITAB} \citep{lu-etal-2023-scitab} contains claims requiring compositional reasoning for verification. We follow the original work to convert it from the setting of fact verification to TQA.

\paragraph{TQA systems for comparison.}
We categorize the recent TQA systems into two groups indicating if they require LLM fine-tuning or not. 
We include the following baselines that require fine-tuning: OmniTab \citep{jiang-etal-2022-omnitab}, TableLlama \cite{zhang2024tablellamaopenlargegeneralist}, ProTrix \citep{Wu2024ProTrixBM}, TAT-LLM \citep{Zhu2024TATLLMAS} and TableLLM \cite{Zhang2024TableLLMET}. 
Except for OmniTab, which is backboned by BART \citep{lewis-etal-2020-bart},
all others build on top of LLaMA 7b \citep{touvron2023llamaopenefficientfoundation}.
Our set of TQA baselines
that do not require fine-tuning includes Dater \citep{Ye2023LargeLM}, Binder \citep{cheng2023binding}, Chain-of-Table \citep{wang2024chain}, ReAcTable \citep{zhang2023reactable}, TabSQLify \citep{nahid2024tabsqlify}, Plan-then-Reason \citep{Wu2024ProTrixBM}, Mix-SC \citep{Liu2023RethinkingTD} and ARC \cite{zhang-etal-2023-crt}.
They all rely on GPT-3.5-turbo.

\paragraph{Experimental settings.}
In MACT, the choice of the planning and coding agent is flexible as no fine-tuning is involved. 
We experiment with the best open-weight LLMs available at the time of writing: Qwen-2 72B \citep{qwen2} (planning agent) and CodeLLaMA-34B (coding agent). 
They are run on int4 and full precision, respectively. 
We further use GPT-3.5-turbo as both the planning and coding agents when comparing our method with other TQA systems that use this model.
$\tau_p$ and $\tau_c$ are set to 0.6 for non-repetitive action generation.
We set the action and code generation size $k$ to 5, following \citet{Liu2023RethinkingTD}. 
The maximum number of iteration $I$ is set to 7, based on empirical results on the development sets.
For the efficiency component, we set $\alpha$ to 1 to ensure high confidence of the model. We explore the effects of different values of $\alpha$ in Section \ref{efficiency_op}.
Two NVIDIA A100 GPUs are used for running MACT.

\section{Results}
\label{sec:results}

We first evaluate the performance of MACT in direct comparison to recent TQA systems using closed-source LLMs.
Second, we examine how our method performs compared to fine-tuned open-weight LLMs.\footnote{We did not run ARC as no code is available. Results for TAT-LLM is only reported on TAT as the model is specially designed for TAT dataset that features both tables and texts as inputs.}
Following prior work, we use exact match (EM) as the evaluation measure. Lastly, we discuss the efficiency of MACT.

\paragraph{MACT outperforms TQA models on three out of four datasets when using GPT-3.5 as the backbone.
}
The upper part of Table~\ref{tab:main_result} shows MACT using GPT-3.5 as the planning and coding agent in comparison to SoTA TQA systems using GPT-3.5 as the backbone LLM.
MACT (GPT-3.5) surpasses the examined TQA models, except for Mix-SC on WTQ. This indicates the effectiveness of our multi-agent strategy compared to single-agent TQA models.
We suspect that the performance gap between our approach and Mix-SC comes from data-specific table-cleaning and answer format controls in Mix-SC. In contrast, MACT does not include any dataset-specific pre- or postprocessing steps to keep it generally applicable to any dataset.
%
\begin{table}[!t]
    \small
    \centering
    \setlength{\tabcolsep}{3pt}
    \begin{tabular}{@{}l l l l l@{} }
         \toprule
          &  \textbf{WTQ} & \textbf{TAT}& \textbf{CRT}& \textbf{SCT} \\
          \midrule
          \multicolumn{5}{c}{\emph{closed-source LLM backbones}} \\
         \midrule
         GPT-3.5 & 45.8& 39.7 & 39.3& 48.9 \\
         Dater & 52.8*& 22.1 &46.8 & 47.1 \\
         Binder & 56.7*& 0.9 & 1.24 &29.1 \\
         Chain-of-Table & 59.9*&20.5 & 33.9 & 27.6 \\
         ReAcTable & 52.4*& 9.26& 29.8 & 32.1\\
         TabSQLify & 64.7*& 13.7 & 42.0 & 50.9 \\
         Plan-then-Reason & 65.2*& 41.2 & 44.9 & 52.5 \\ 
         Mix-SC & \textbf{73.6*} & 54.3 & 48.6 & 49.3 \\
         ARC & - & - & 56.3* & - \\
         \textbf{MACT} & 
       70.4 & \textbf{64.5} & \textbf{57.4} & \textbf{55.8} \\
         \midrule
                   \multicolumn{5}{c}{\emph{open-weight LLM backbones}} \\
         \midrule
         Qwen (Qw-72b) &60.6 & 53.6& 55.9& 55.0\\ 
         CodeLLaMA (CL-34b)& 55.0& 29.5& 49.7& 9.5 \\ 
         \texttt{SC}(Qw-72b+CL-34b) &69.0 &56.7& 61.4 & 54.4\\
      MACT (Qw-72b+Qw-72b) & 68.6&66.3&59.8 & 57.3\\ 
      MACT (CL-34b+CL-34b) & 55.2&54.1&43.5 &45.0 \\   
                  MACT (Qw-72b+CL-34b)& \textbf{72.6} & \textbf{66.2} & \textbf{64.4} & \textbf{59.8} \\ \midrule
           \textit{GPT-4} & \textit{72.9}$^\dag$  &\textit{80.8}$^\dag$&\textit{58.7}$^\dag$ & \textit{63.2}$^\dag$\\
         \textit{Humans (Crowdsourcing)} &- & \textit{84.1}$^\dag$& - & \textit{84.7}$^\dag$\\
    \bottomrule
         
    \end{tabular}
    \caption{Exact Match results of \textbf{models without fine-tuning.} SCT refers to SCITAB. The models are grouped by the LLM they use as their backbone (top: GPT-3.5; middle: open-weight LLMs as indicated in parentheses). Performances marked with * are taken from the original paper. Performances marked with \dag are taken from \citet{Wu2024ProTrixBM}, \citet{Zhu2024TATLLMAS}, \citet{zhang-etal-2023-crt} and \citet{lu-etal-2023-scitab} for each dataset. We bold the best performances in each group.}.
    \label{tab:main_result}
\end{table}

\paragraph{MACT outperforms out-of-the-box open-weight LLMs across datasets, demonstrating the effectiveness of specialized agents.}
The middle part of Table \ref{tab:main_result} provides the results of MACT using specialized agents (MACT (Qw+CL): Qwen-2 as planning agent, CodeLLaMA as coding agent).
As baselines, we compare a setting without a specialized coding agent (MACT (Qw+Qw)) and a setting without a general planning agent (MACT(CL+CL)). 
As further baselines, we use the two LLMs on their own as well in combination, with Chain-of-thought prompting \citep{wei2023chainofthoughtpromptingelicitsreasoning}.
For combination, the single models are prompted five times and combined using \texttt{SC}, as in \citet{Liu2023RethinkingTD}.
This is a direct multi-agent baseline without collaboration and tool use.
Both \texttt{SC}(Qw+CL) and MACT (Qw+CL) achieve higher EM scores than individually prompting Qwen and CodeLLaMA, demonstrating the positive effect of using multiple agents for planning and coding.
Importantly, MACT (Qw+CL) outperforms \texttt{SC}(Qw+CL) by approximately 6 EM points on average across all datasets, highlighting the superiority of our collaboration technique over simply taking the most frequent predictions from two independent agents.
We also find that having an expert coding agent for code generation (MACT (Qw+Qw) vs.\ MACT (Qw+CL)) improves performance considerably.
%
%

\paragraph{MACT with open-weight models delivers comparable performance as closed-source systems.}
As shown in Table \ref{tab:main_result}, by comparing MACT (Qw+CL) with TQA systems that rely on closed-source LLMs (in the upper part of Table \ref{tab:main_result}), we find that our model outperforms the examined TQA systems for three out of four datasets.
Our multi-agent TQA system is more cost-efficient and straightforward to replicate, while delivering superior performance compared to closed-source TQA models.
In addition, we show two upper-bounds in the bottom part of Table \ref{tab:main_result}: The performance of human annotators and directly prompting GPT-4 with the table and question.
As shown in Table \ref{tab:main_result}, GPT-4 has an advantage on TAT and SCITAB.
On WTQ, we observe comparable performances between MACT (QW+CL) and GPT-4.
On CRT, our method even outperforms GPT-4 by 5.7\%.
CRT is the most complex dataset, requiring multi-step and multi-category reasoning, which direct inference with GPT-4 cannot generally solve.
Our step-wise collaborative planning setting is well-suited to such settings.
In contrast, there is a large gap between MACT and human performance in SCITAB. 
SCITAB collects data from scientific papers, in which abbreviations and domain-specific terms are common. These can pose challenges to current systems and models.
In TAT, MACT often finds the correct answer but struggles to output it in the correct format (see Sec. \ref{error_analysis}).


\paragraph{MACT generalizes better across datasets than fine-tuned TQA systems.}
Table \ref{tab:main_result_ft} compares our framework with prior fine-tuned TQA models.
We present results for MACT with different planning and coding agents: Our standard setting (LlaMA-7b+CS-7b) as well as with a stronger planner (Qw-7b+CS-7b).
In general, for fine-tuned TQA models, their performance on the dataset used for fine-tuning is rather high while they suffer from a considerable drop in EM when tested on other datasets. This observation is in line with the findings by \citet{zhang2024tablellamaopenlargegeneralist} and \citet{huang2024limitationsfinetunedjudgemodels}. 
In contrast, MACT does not use fine-tuned models and can, thus, be applied to any dataset with a good generalization performance. 
MACT demonstrates comparable results to Protrix when using LlaMA-7b as the planning agent, though it has not been fine-tuned. As expected, using a better planning agent leads to better results. This also shows the robustness of MACT in terms of backbone models.

\begin{table}[!t]
    \small
    \centering
    \begin{tabular}{@{}l@{\textcolor{white}{c}} l l l l@{}}
         \toprule
          & \textbf{WTQ} & \textbf{TAT}& \textbf{CRT}& \textbf{SCT} \\
          \midrule
         OmniTab (BART 406m)& 62.3* & 17.1 & 20.6 & 29.1 \\ 
         TableLlama (LlaMA-7b) & 29.9& 17.4 & 26.9 & 38.6 \\ 
         Protrix (LlaMA-7b)& 48.9*& 26.8& 40.2 &42.4 \\
         TAT-LLM (LlaMA-7b) & -& \textbf{69.6*} & - & - \\
         \midrule
         
        MACT (LlaMA-7b+CS-7b) &38.1 & 28.3& 40.0&41.1 \\
         MACT (Qw-7b+CS-7b) & \textbf{58.4}&61.9 & \textbf{46.4}&\textbf{45.9} \\
    \bottomrule
         
    \end{tabular}
    \caption{Exact Match Results of MACT using different LLM agents in \textbf{comparison to fine-tuned TQA models}. 
 Performances marked with * refer to the in-domain setting (where fine-tuning took place). SCT refers to SCITAB. CS refers to the deepseek-coder model.
    }
    \label{tab:main_result_ft}
\end{table}

%

\paragraph{MACT adapts computational cost to instance complexity.}
Table \ref{tab:computational_cost} compares MACT with other approaches in terms of the total number of 
LLM calls 
for each instance. 
For Binder and Dater, \texttt{SC} is performed a fixed number of times regardless of problem complexity. This results in a high number of LLM calls 
per instance, making them inefficient. 
In contrast, MACT provides flexibility in generation, as the number of iterations depends on the problem's complexity. For instance, most questions can be solved within three steps for WTQ (see our analysis in \ref{iteration_num}). This results in a total of at most 25 LLM calls 
\footnote{2 steps involve action and execution generation, with each five times, plus last step five times of action generation: 2*(5+5)+5.} for each instance. If we incorporate the efficiency optimization module, which potentially saves up to one-third of the iterations (see Section \ref{efficiency_op}), the total number of LLM calls 
per instance is even lower (approximately 15), making MACT comparable to other approaches in terms of efficiency. 
The iterative nature of MACT can lead to a higher upper-bound of LLM calls. 
However, it also allows for solving more complex problems, making the approach more tailored to real-life requirements.

\begin{table}[!t]
    \small
    \centering
    \begin{tabular}{@{}l@{\textcolor{white}{c}} c  @{}}
         \toprule
          & \textbf{Number of LLM calls per instance} \\
          \midrule
         Binder& 50  \\
        Dater& 100  \\ 
        Chain-of-Table &  1-25 \\
        ReAcTable& 15-125\\
         Mix-SC& 10-30 \\
       MACT& 5-65\\
    \bottomrule
         
    \end{tabular}
    \caption{Number of LLM calls for different approaches. We show lower and upper-bounds if not deterministic.}
    \label{tab:computational_cost}
\end{table}

\begin{table}[t]
\centering
\footnotesize
  \begin{tabular}{lcccc}
    \toprule
      & \textbf{WTQ} &  \textbf{TAT} &  \textbf{CRT}&  \textbf{SCITAB}\\


    \midrule
  MACT (Qw-72b) & \textbf{72.6} & \textbf{66.2} & \textbf{64.4} & \textbf{59.8} \\\cmidrule{2-5}
 \hspace{2mm} w/o $T_{srh}$ & \makecell{72.0} &\makecell{66.2}&\makecell{64.6} & \makecell{59.6}\\
     \hspace{2mm} w/o $T_{srh}+T_{cal}$ & \makecell{71.3}& \makecell{62.8} & \makecell{63.9}& \makecell{58.2} \\  

     \hspace{2mm} w/o $T$+$M_c$  & \makecell{67.1}& \makecell{61.2} & \makecell{60.4} &\makecell{57.9}\\
    \bottomrule
  \end{tabular}
\caption{Ablation study. $T_{srh}$ and $T_{cal}$ refer to the Wikipedia search tool and the calculator tool. $T$ includes the above two tools and a Python interpreter. $M_c$ is the coding agent.}
\label{tab:ablation}
\end{table}

\section{Analysis}
\label{analysis}
We conduct various analyses of our framework to back up our claims and contributions.
Unless mentioned otherwise, all analyses are performed using Qwen-2 72B as $M_p$,  CodeLlama-34B as $M_c$, the number of action generation $k=5$, and selection model $f_s = \texttt{SC}$. To explicitly analyze the effects of multi-agent collaboration with tool use, we do not use efficiency optimization, which means all instances undergo the iterative collaboration between $M_p$ and $M_c$ with tool use. 
Further analysis to support our choices of the sampling size $k$ and the maximum number of iterations $I$ are in \ref{sample_size} and \ref{iteration_num}.
A case study can be found in \ref{case_study}.

\begin{figure}[t]
    \centering
    \includegraphics[width=1\columnwidth]{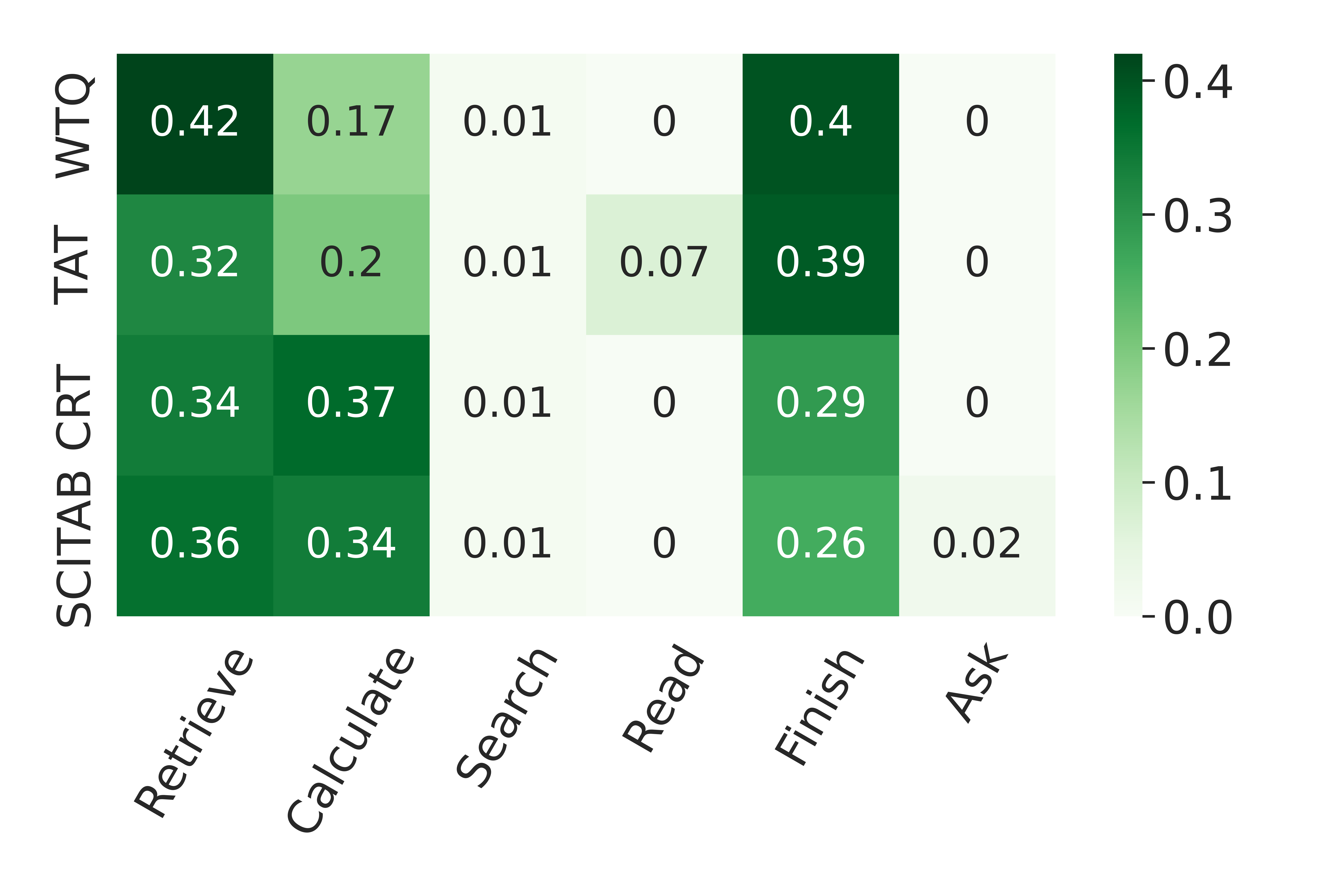}
    \caption{Distribution of action intents by dataset.}
    \label{fig:intent}
\end{figure}

\paragraph{Effect of Multi-Agent-Collaboration with Tool Use.}
We explore the effectiveness of specialized agents and tool use in MACT by conducting an ablation study with three scenarios: 
ablating only $T_{srh}$ (Wikipedia search API), 
ablating $T_{srh}$ and $T_{cal}$ (calculator), and further ablating the coding agent $M_c$ with a Python interpreter.
In cases where $M_c$ or/and tools are ablated, the most frequent estimated observations from $M_p$ are used as the final observations.
Our results in Table \ref{tab:ablation} show that both the tools and the coding agent contribute to the performance of the framework. 
Nevertheless, they contribute differently to the final performance. 
For instance, ablating the search tool barely influences the results whereas there are large performance drops when further ablating the coding agent and the Python interpreter.
We find that the search tool is barely used whereas the coding agent is called in almost every query.
Since Wikipedia is a common pre-training corpus for LLMs, most information might have already been encoded in the LLM.
Nevertheless, the search tool can still be helpful given LLMs are known to suffer from hallucinations and the knowledge encoded might not be updated in time.
For more specialized domains and sources, the search tool may be crucial. 
We further observe that the ablation affects WTQ and TAT more than CRT and SCITAB.  
This might be attributed to dataset features: CRT contains many yes-no questions and SCITAB has been converted from a ternary classification dataset. 
Thus, chances for guessing the correct final answers are higher than in datasets with a more diverse answer distribution, such as WTQ and TAT. 
By evaluating our framework on instances from CRT that have answers other than yes/no, we find a performance drop of 8.23 when ablating both tools and the coding agent.

\begin{figure}[t]
    \centering
    \includegraphics[width=1.\columnwidth]{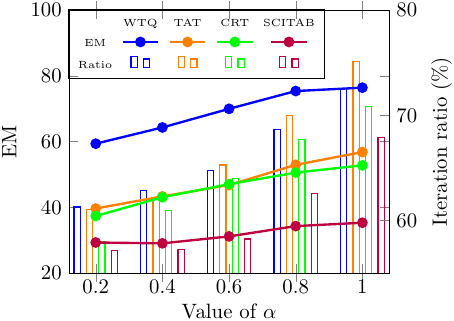}
    \caption{EM (line chart) and iteration ratio (bar chart) against different $\alpha$. The iteration ratio is calculated by dividing the number of iterations when using efficiency optimization with a specific $\alpha$ by the number of iterations when not using it (without shortcuts). }
    \label{fig:efficiency}
\end{figure}

\paragraph{Analysis of Intent Distribution.}
We report the distribution of the intents of the selected actions for each dataset in Figure \ref{fig:intent}. 
We observe that \textit{Retrieve} and \textit{Calculate} are the most frequent intents, along with \textit{Finish}.
This indicates that our proposed second agent, the coding agent $M_c$ is used frequently.
Different datasets also require different intents. 
In particular, the framework needs to use the intent \textit{Read} for solving instances in TAT, where textual descriptions are given while this is not the case for the other datasets.
We notice that the \textit{Search} intent is used only few times across datasets.
This might be because most instances were designed to be solved using the given table and text information.
However, when looking into the individual cases where \textit{Search} is used, we still find it useful.
For instance, one question from WTQ asks about the number of athletes from American but no information about nationality is given in the table.
In this case, \textit{Search} assists with answering the question by adding the nationality for each athlete from Wikipedia.
Though the intents \textit{Read}, \textit{Search} and \textit{Ask} are less used compared to others, we still incorporate them to adapt to various use cases that might occur in real-life use cases.

\paragraph{Effect of Efficiency Optimization.}
\label{efficiency_op}

To investigate the trade-off between efficiency and accuracy, we plot the model performance against $\alpha\in \{0.2, 0.4, 0.6, 0.8, 1\}$ in Figure \ref{fig:efficiency}.
We also plot the ratio of the total number of iterations taken to terminate 
from each tested $\alpha$ to the total number of iterations taken when not using the optimization, i.e.,  letting the planning agent decide via the \textit{Finish} action when to stop the execution.
The best performance is reached for $\alpha=1$, i.e., when requiring all estimated results to agree with each other to stop the iteration.
For SCITAB, for instance, we save approximately 40\% of the iterations when setting $\alpha$ to 1 without losing performance compared to not using the optimization component (59.8\% vs.\ 59.7\%).
On average, adding the efficiency optimization module saves up to 33\% of iterations. 
This shows the effectiveness of the optimization and that users can individually tune the desired trade-off of performance and computation time.

\paragraph{Error Analysis.}
\label{error_analysis}
We randomly sample 50 instances that MACT fails per dataset and conduct an error analysis. 
About half of the errors come from invalid or wrong code generated by the coding agent $M_{c}$. Either $M_c$ fails to make sense of instructions 
or of complex table structure. 
The second error type can be attributed to evaluation.
We find that about one-third of failures come from strict evaluation metrics (EM). This influences the performance of MACT particularly on the TAT dataset, as it features long text strings as answers.
The evaluation challenge has been discussed in many previous works \cite{Wu2024ProTrixBM,Li2024GraphOTTEREL}. To estimate the upper-bound of our method, we use GPT-4 as evaluator to determine if a predicted answer is semantically the same as the reference answer. 
This results in an accuracy of 87.8\% on the TAT dataset, compared to 66.2\% with EM.
The remaining error cases can be largely attributed to the failure of the planning agent 
in decomposing questions correctly. For instance, one question asks for the score range (min-max) of the top 10 finishers. Apart from retrieving the min and max scores of the top 10 finishers, the planner continues to generate the action: \texttt{Calculate[Calculate the range of the scores in the observation 1.]}. This leads to a wrong prediction.

\section{Conclusions}
We have proposed MACT, a multi-agent collaboration with tool \anne{use} for table question answering. 
Unlike previous work,  MACT neither requires fine-tuning nor does it depend on closed-source models.
In our experiments, our framework demonstrates good generalizability across different benchmark datasets and outperforms a number of state-of-the-art approaches, including closed-source commercial models and fine-tuned models. 
To boost efficiency, we introduce an efficiency optimization module that saves up to 33\% of the iterations in our analysis.
In our experiments and analyses, we show that multi-agent collaboration with tools is an effective approach for table question answering.

\section{Limitations}
MACT is evaluated mainly with single table settings due to the scarcity of datasets featuring multi-table complex reasoning. Though the framework can be extended easily to deal with multiple tables by concatenating them in the inputs, it is still not clear how effective our approach will be in a multi-table setting. Secondly, we only study TQA in the context of English, while there exist many multi-lingual TQA benchmarks and challenges. 



\section*{Acknowledgements}
This work was partially supported by the EU Project SMARTY (GA 101140087).

\bibliography{anthology,custom}
\bibliographystyle{acl_natbib}

\appendix

\clearpage
\section{Appendix}
\label{appendix}

\subsection{Datasets}
\label{datasets}
Table \ref{datasets} shows their statistics and characteristics.
WTQ \cite{Pasupat2015CompositionalSP}, TAT \citep{zhu-etal-2021-tat}, CRT \cite{zhang-etal-2023-crt} and SCITAB \citep{lu-etal-2023-scitab} are publicly available under the
licenses of CC-BY-SA-4.05, MIT and MIT, respectively. These licenses all permit us to compose, modify, publish, and distribute additional annotations upon the original dataset.

\begin{table}[!h]
\small
\centering
  \begin{tabular}{@{}llccl@{}}
    \toprule
    \textbf{Dataset}&
      \textbf{\#Test} &  \textbf{M-step} &  \textbf{M-category}&  \textbf{Domain}\\
    \midrule
    WTQ & 4,344 &\xmark&\xmark&General\\
    TAT &1,663&\cmark&\xmark&Financial\\
    CRT & 728&\cmark&\cmark&General\\
    SCITAB & 1,162 &\cmark&\cmark&Scientific\\
    \bottomrule
  \end{tabular}
\caption{
We use four datasets that vary in reasoning complexity (M-step: multi-step, M-category: multi-category reasoning) and domain. \#Test refers to the number of test instances. 
}
\end{table}

\subsection{Effect of Sampling Size}
\label{sample_size}
\begin{figure}[!hbt]
    \centering
    \includegraphics[width=1.\columnwidth]{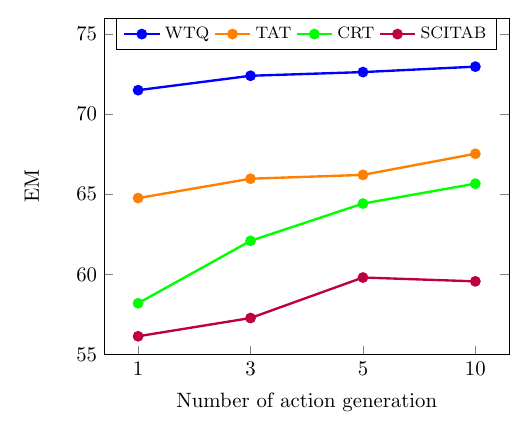}
    \caption{EM against different action generation size.  }
    \label{fig:sampling_size}
\end{figure}

Figure \ref{fig:sampling_size} shows the effect of the number  of generated actions $k$ on the results. 
 Generally, we find that a larger $k$ results in better performance. However, the performance gain is small when increasing $k$ from 5 to 10.
 We even observe a slight performance drop for SCITAB when increasing $k$ from 5 to 10. Based on these observations, we argue $k=5$ is a good choice for the number of generated action in MACT.

\subsection{Analysis of Iteration Number Distribution}
\label{iteration_num}
\begin{figure}
    \centering
    \includegraphics[width=0.9\columnwidth]{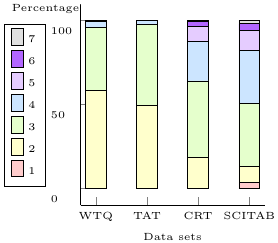}
    \caption{The distribution of number of iterations for each dataset.}
    \label{fig:iteration}
\end{figure}

We analyze the distribution of numbers of iterations for each dataset in Figure \ref{fig:iteration}.  
Most of instances can be solved within seven iterations. 
Dataset-wise, CRT and SCITAB seem to require more iterations than WTQ and TAT, indicating their difficulties in terms of multi-step reasoning.

\subsection{Choice of Selection Model.}
\label{selection_model}
In MACT, we use \texttt{SC} as the action selection model (see Section \textbf{Action Selection}). We now provide results for alternative selection models that have been introduced by prior work.
In particular, we compare to LLM-based selection \citep{yao2023treethoughtsdeliberateproblem}, log probability \citep{zhang2023reactable}, roll-out \citep{hao-etal-2023-reasoning} and a combination of all strategies \citep{hao-etal-2023-reasoning}. 
In the \texttt{LLM} strategy, an LLM evaluator is utilized to select the best action. For consistency with our other results,
we use Qwen-72b as the evaluator.
We use the same prompts as the original work \cite{yao2023treethoughtsdeliberateproblem}. 
Log probability (\texttt{LOG\_P}) has been widely used to assist sub-path selection \citep{zhang2023reactable,hao-etal-2023-reasoning}. 
However, it can only be used for open-weight LLMs, as it requires access to log probabilities.
\texttt{ROLL\_OUT} 
estimates future answers by rolling out the current reasoning path and selects the action that leads to the most frequent future answer.
For the \texttt{COMBINED} method, we use majority voting among all individual selection models.
\begin{table}[!t]
\centering
\small
  \begin{tabular}{@{}lcccc@{}}
    \toprule
      & \textbf{WTQ} &  \textbf{TAT} &  \textbf{CRT}&  \textbf{SCITAB}\\
    \midrule
    \textsc{sc}& 72.6 (2) & 66.2 (2) & \textbf{64.4} (1) &\textbf{59.8} (1) \\
    \textsc{llm} &70.7 (4) &66.2 (2) &58.4 (5) &56.8 (4)\\
    \textsc{log\_p} &70.1 (5) &64.9 (5) & 61.4 (3)&57.2 (3) \\
    \textsc{roll\_out} &71.9 (3) &65.9 (4) &60.2 (4) & 55.5 (5)\\
    \textsc{combined} &\textbf{72.7} (1) &\textbf{66.6} (1) & 62.6 (2) &57.4 (2)\\
    \bottomrule
  \end{tabular}
\caption{Results using different selection models. We put the relative ranking of the models per dataset in parentheses.}
\label{tab:result_selection}
\end{table}
The results in Table \ref{tab:result_selection} show that for WTQ and TAT, \texttt{SC} and \texttt{COMBINED} lead to the best performance.
For CRT and SCITAB, \texttt{SC} outperforms \texttt{COMBINED}, caused by the comparably poorer performance of \texttt{LLM}, \texttt{LOG\_P} and \texttt{ROLL\_OUT} on these datasets. 
\texttt{SC} is more efficient than \texttt{COMBINED} as the latter requires running all selection models, including the computationally expensive LLM.
Overall, this analysis confirms that \texttt{SC} as selection model is a good choice. 

\subsection{Case Study}
\begin{figure*}[!hbt]
    \centering
    \includegraphics[width=1.\textwidth]{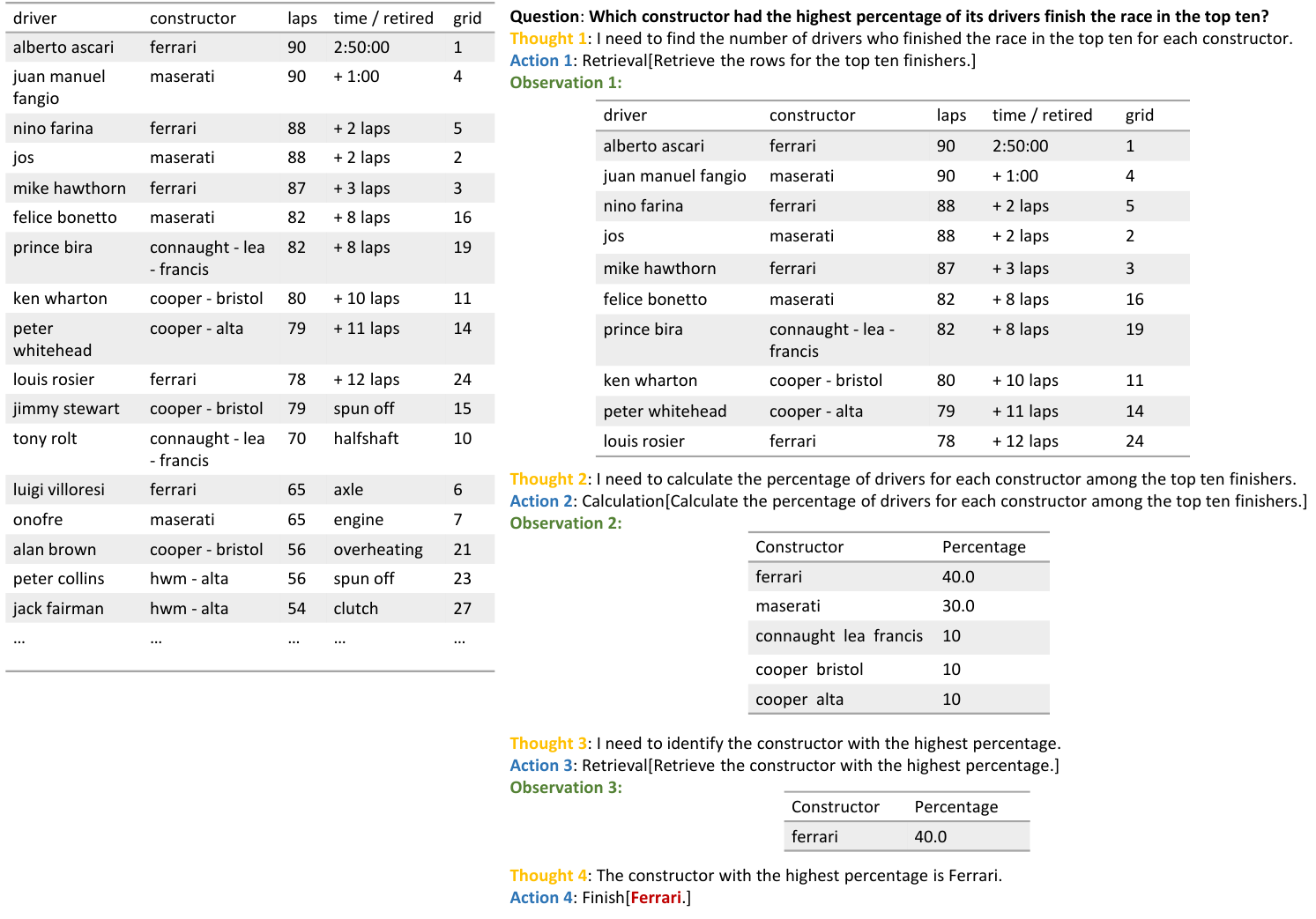}
    \caption{An instance selected from CRT featuring complex reasoning.}
    \label{fig:case1}
\end{figure*}

\begin{figure*}[!hbt]
    \centering
    \includegraphics[width=1.\textwidth]{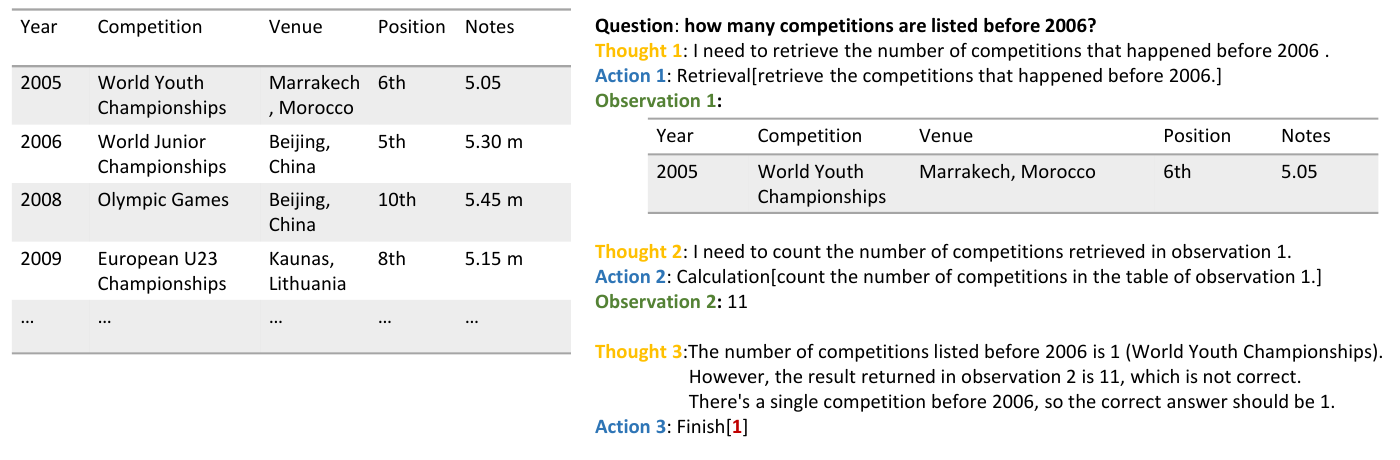}
    \caption{An instance selected from WTQ. We find the planning agent can perform self-correct given previous reasoning traces.}
    \label{fig:case2}
\end{figure*}

We present two reasoning traces selected from CRT and WTQ in Figure \ref{fig:case1} and \ref{fig:case2}, respectively. Figure \ref{fig:case1} presents a case involving complex reasoning, where models need to identify top ten finishers, calculate percentage of drivers for each constructor and return the constructor with the highest percentage. In Figure \ref{fig:case2}, we observe the planning agent has the ability to self-correct in Thought 3.

\label{case_study}

\subsection{MACT Prompts}
\label{prompts}
\begin{figure*}[!htb]
    \centering
    \includegraphics[width=1.\textwidth]{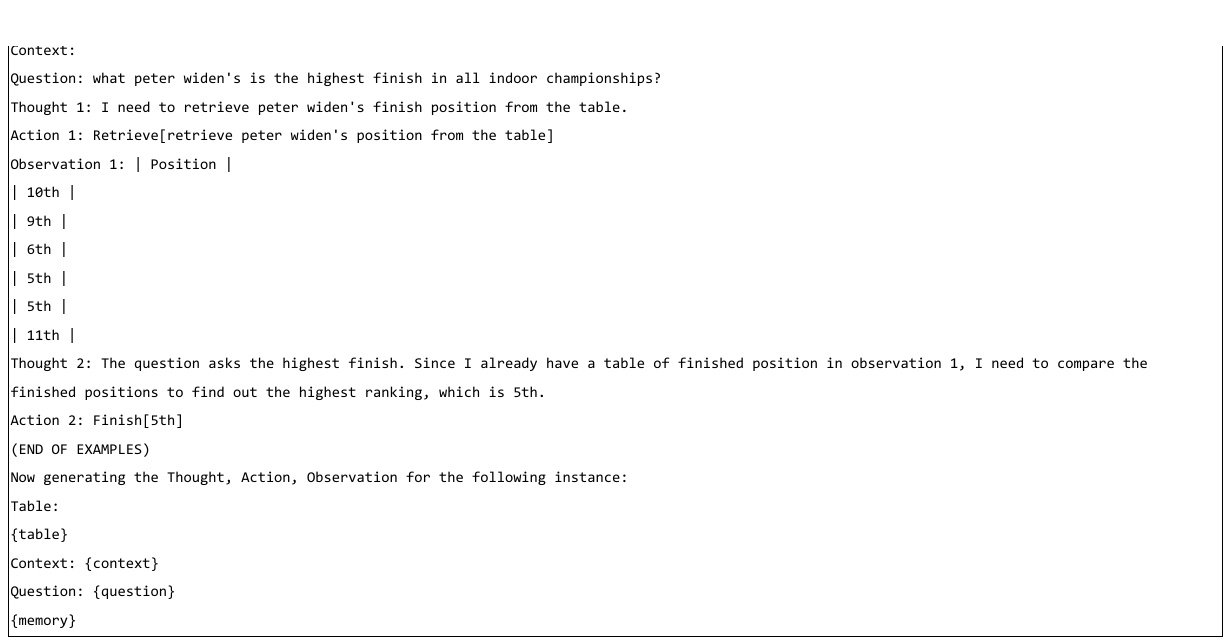}
    \caption{MACT:  Planning agent prompt for WTQ.}
    \label{wtq_prompts}
\end{figure*}

\begin{figure*}[!ht]
    \centering
    \includegraphics[width=1.\textwidth]{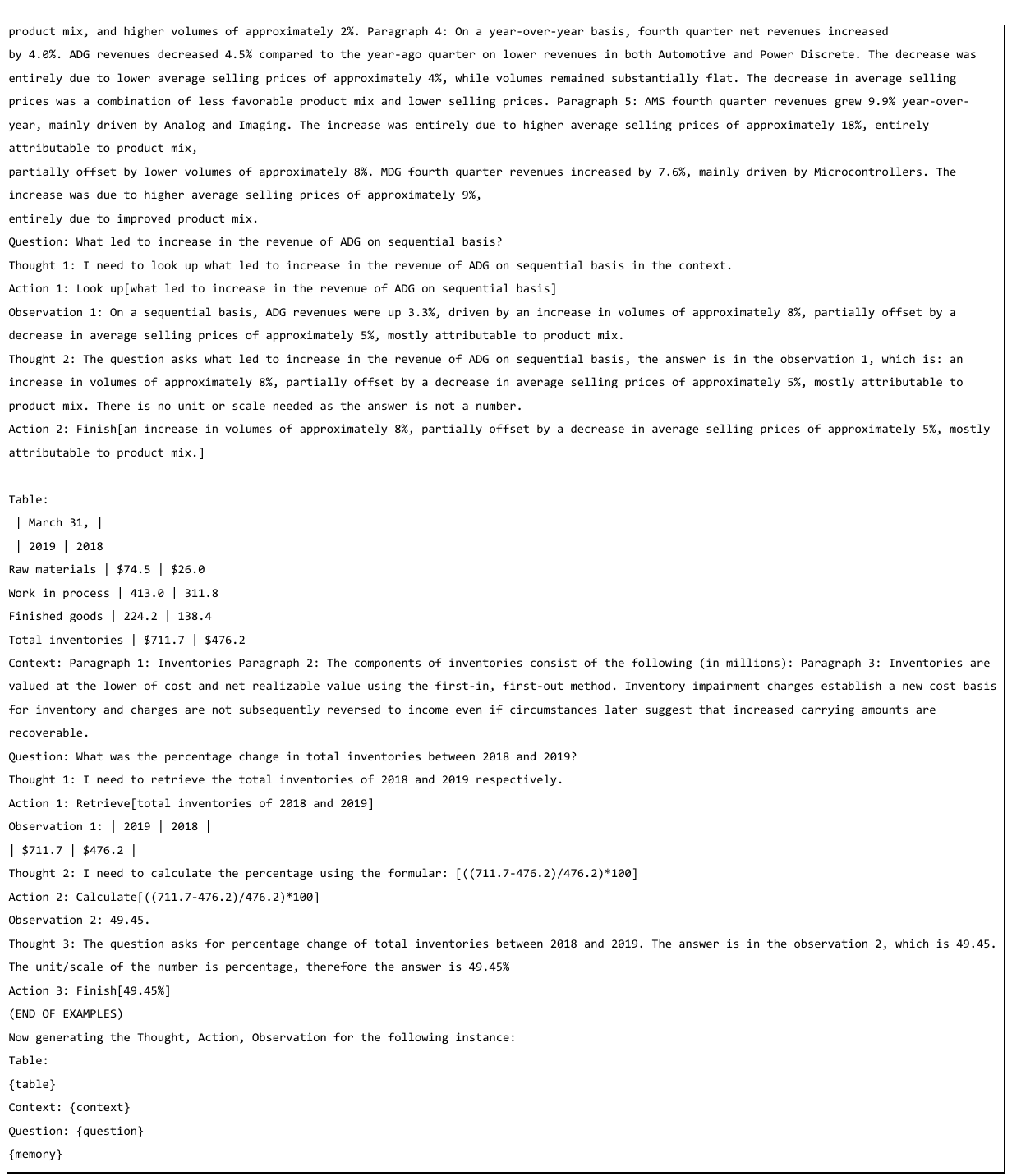}
    \caption{MACT:  Planning agent prompt for TAT.}
    \label{tat_prompts}
\end{figure*}

\begin{figure*}[!ht]
    \centering
    \includegraphics[width=1.\textwidth]{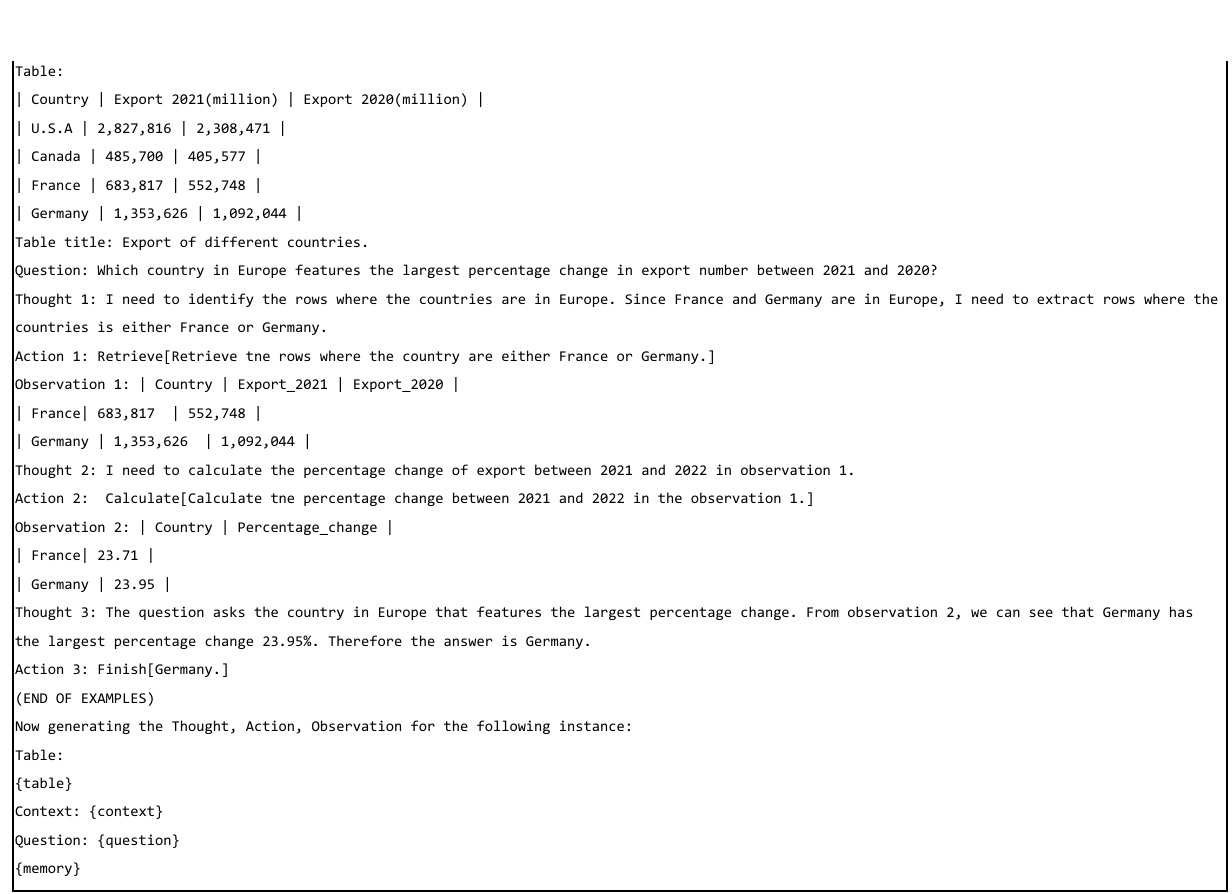}
    \caption{MACT:  Planning agent prompt for CRT.}
    \label{crt_prompts}
\end{figure*}

\begin{figure*}[!ht]
    \centering
    \includegraphics[width=1.\textwidth]{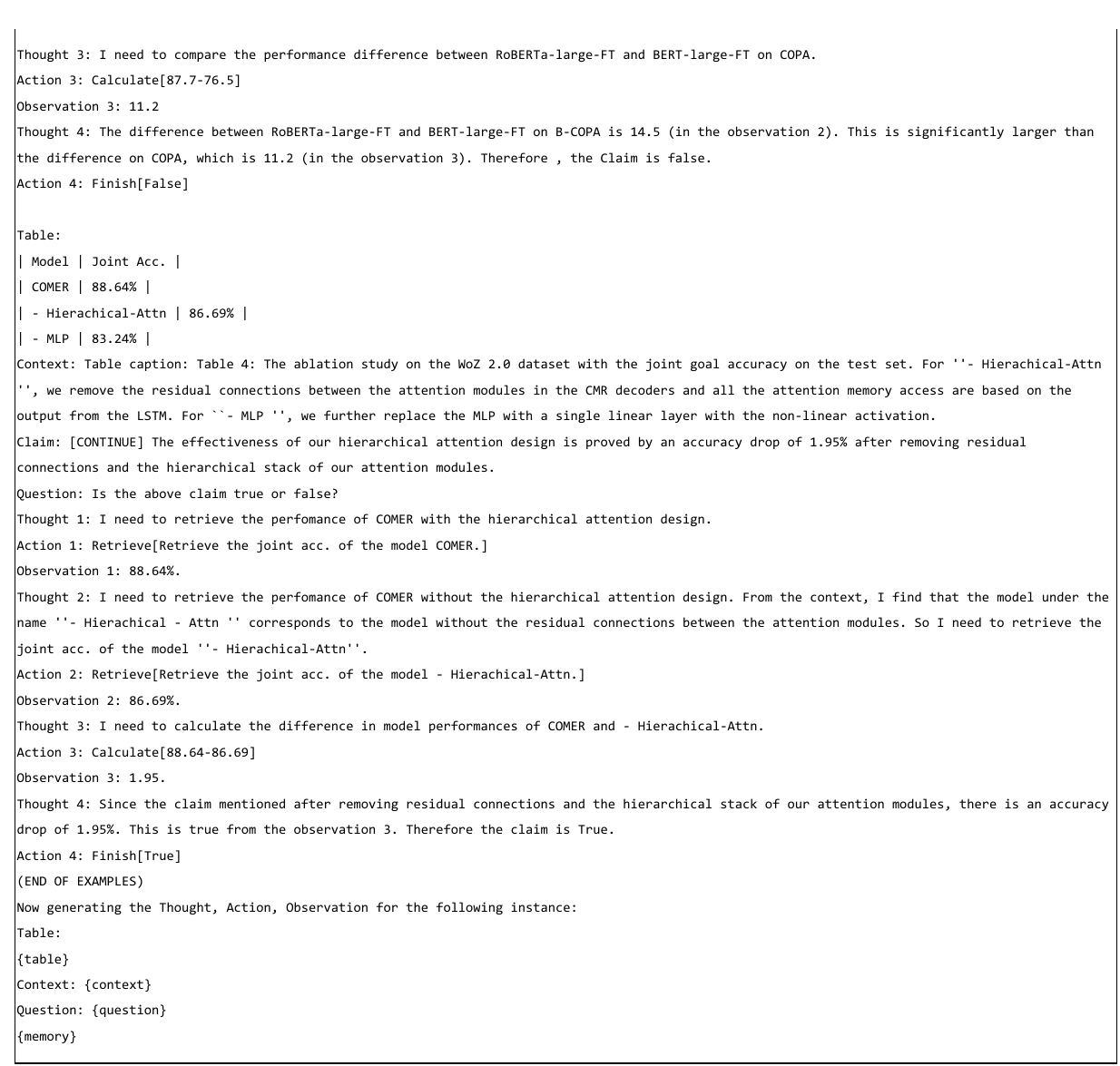}
    \caption{MACT: Planning agent prompt for SCITAB.}
    \label{scitab_prompts}
\end{figure*}

\begin{figure*}[!ht]
    \centering
    \includegraphics[width=1.\textwidth]{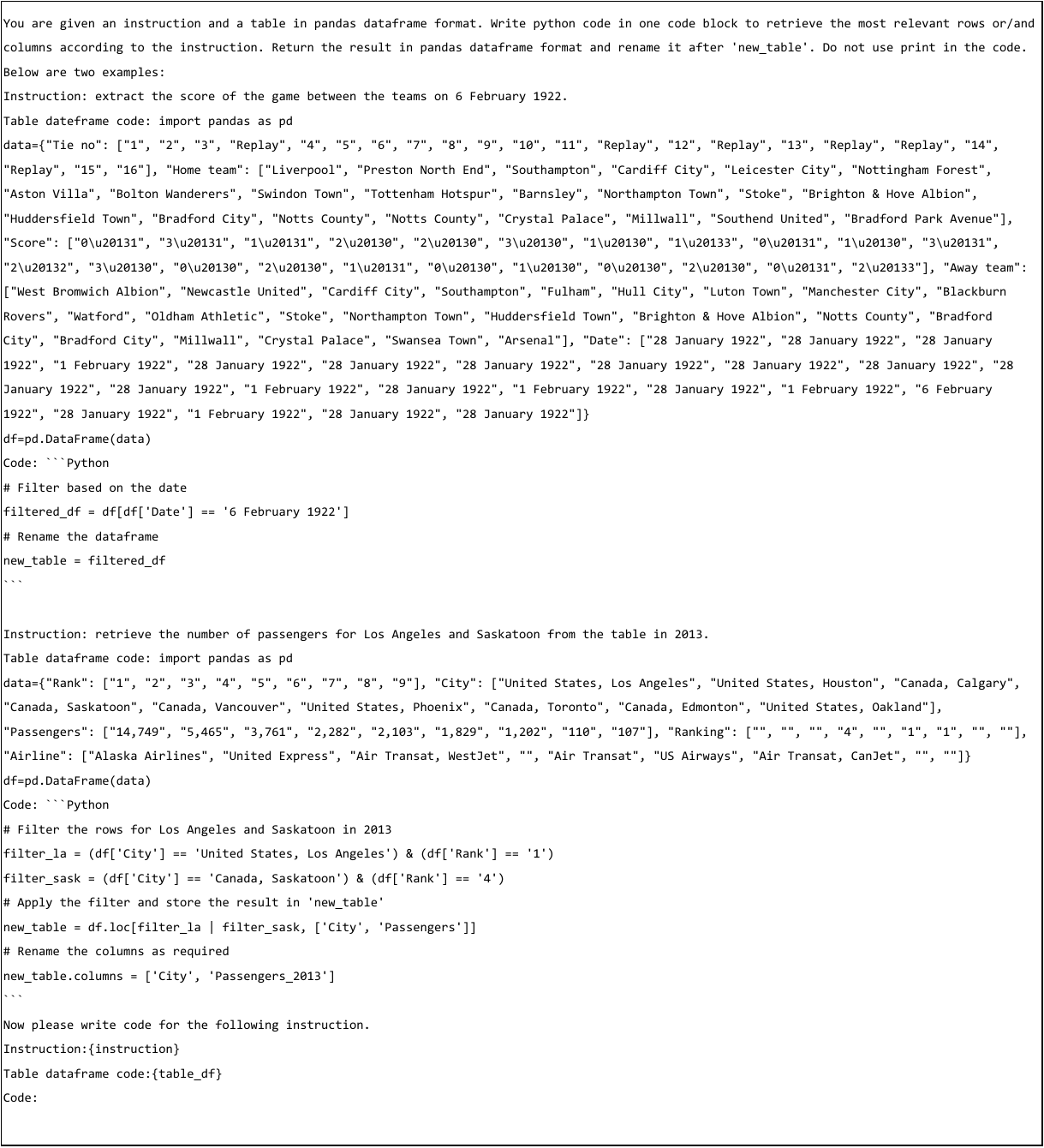}
    \caption{MACT: Coding agent prompt for retrieval.}
    \label{prompt_coding_retrieval}
\end{figure*}

\begin{figure*}[!ht]
    \centering
    \includegraphics[width=1.\textwidth]{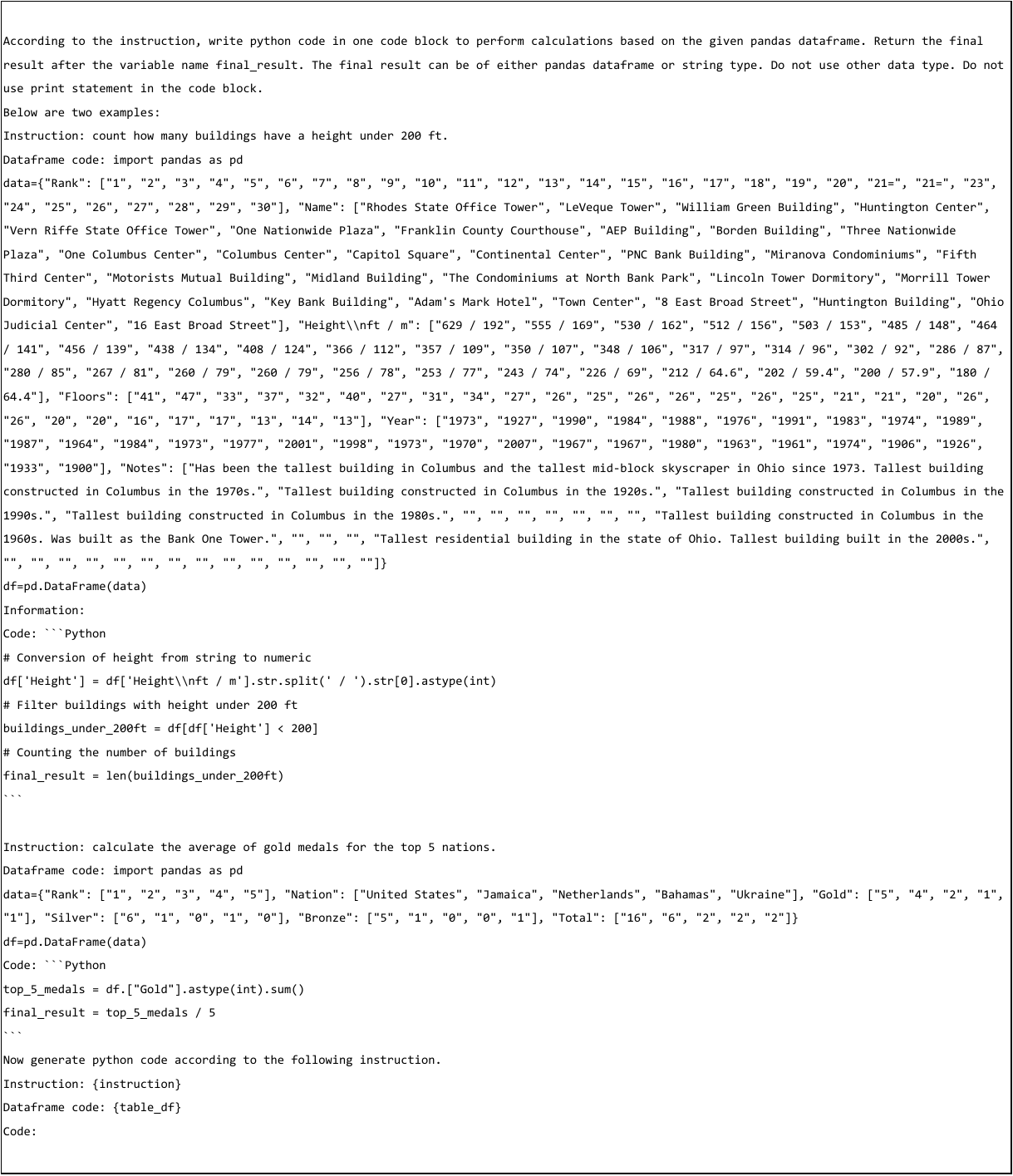}
    \caption{MACT: Coding agent prompt for calculation.}
    \label{prompt_coding_calculation}
\end{figure*}

We provide the prompts used for the planning agent for the examined datasets (WTQ, TAT, CRT and SCITAB) in Figures \ref{wtq_prompts}, \ref{tat_prompts}, \ref{crt_prompts}, and \ref{scitab_prompts}. Figure \ref{prompt_coding_retrieval} and Figure \ref{prompt_coding_calculation} show the prompts used for the coding agent for the action intents \textit{Retrieval} and \textit{Calculation}, respectively. 

\begin{figure*}[!ht]
    \centering
    \includegraphics[width=1.\textwidth]{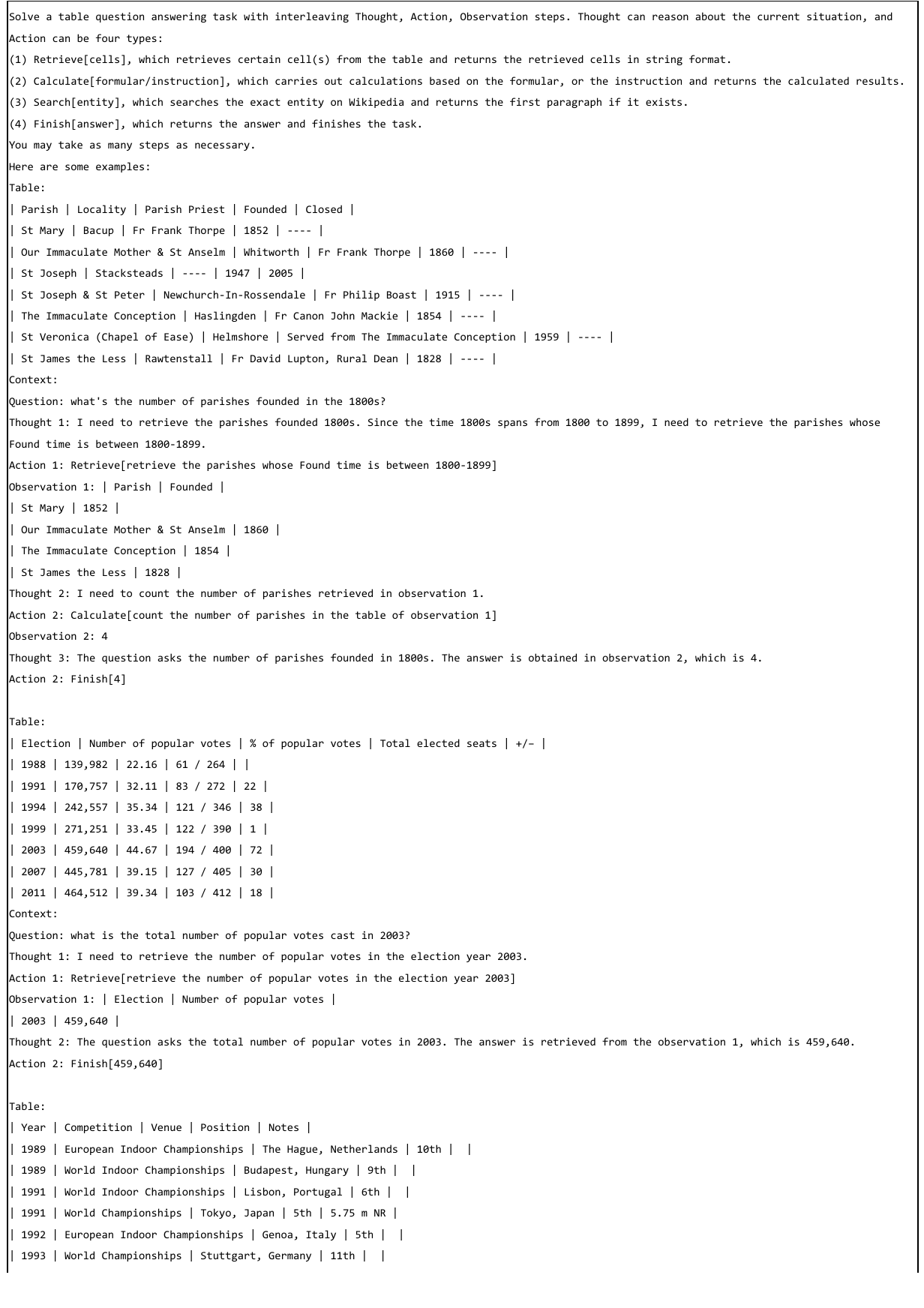}
\end{figure*}

\begin{figure*}[!ht]
    \centering
    \includegraphics[width=1.\textwidth]{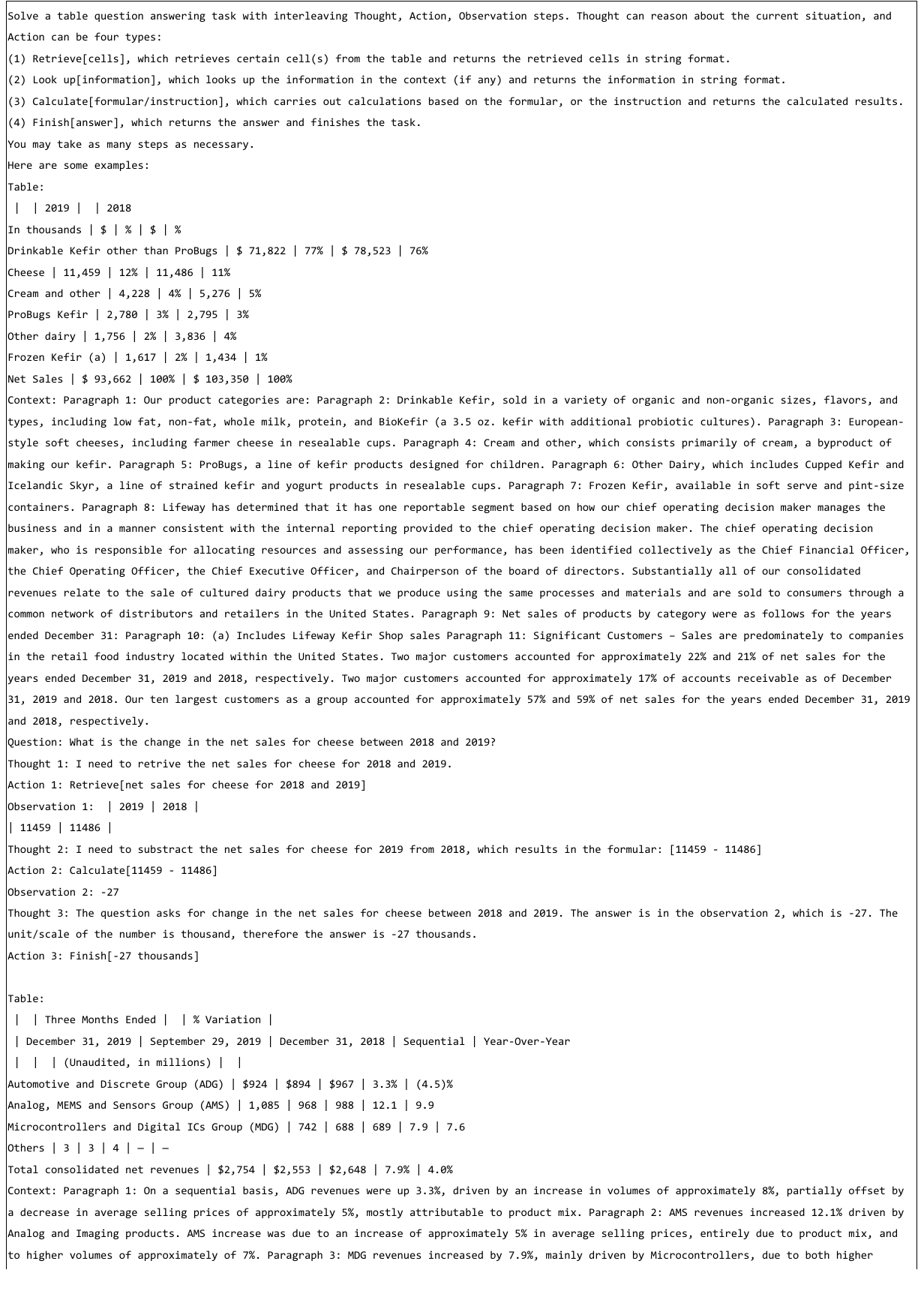}
\end{figure*}

\begin{figure*}[!ht]
    \centering
    \includegraphics[width=1.\textwidth]{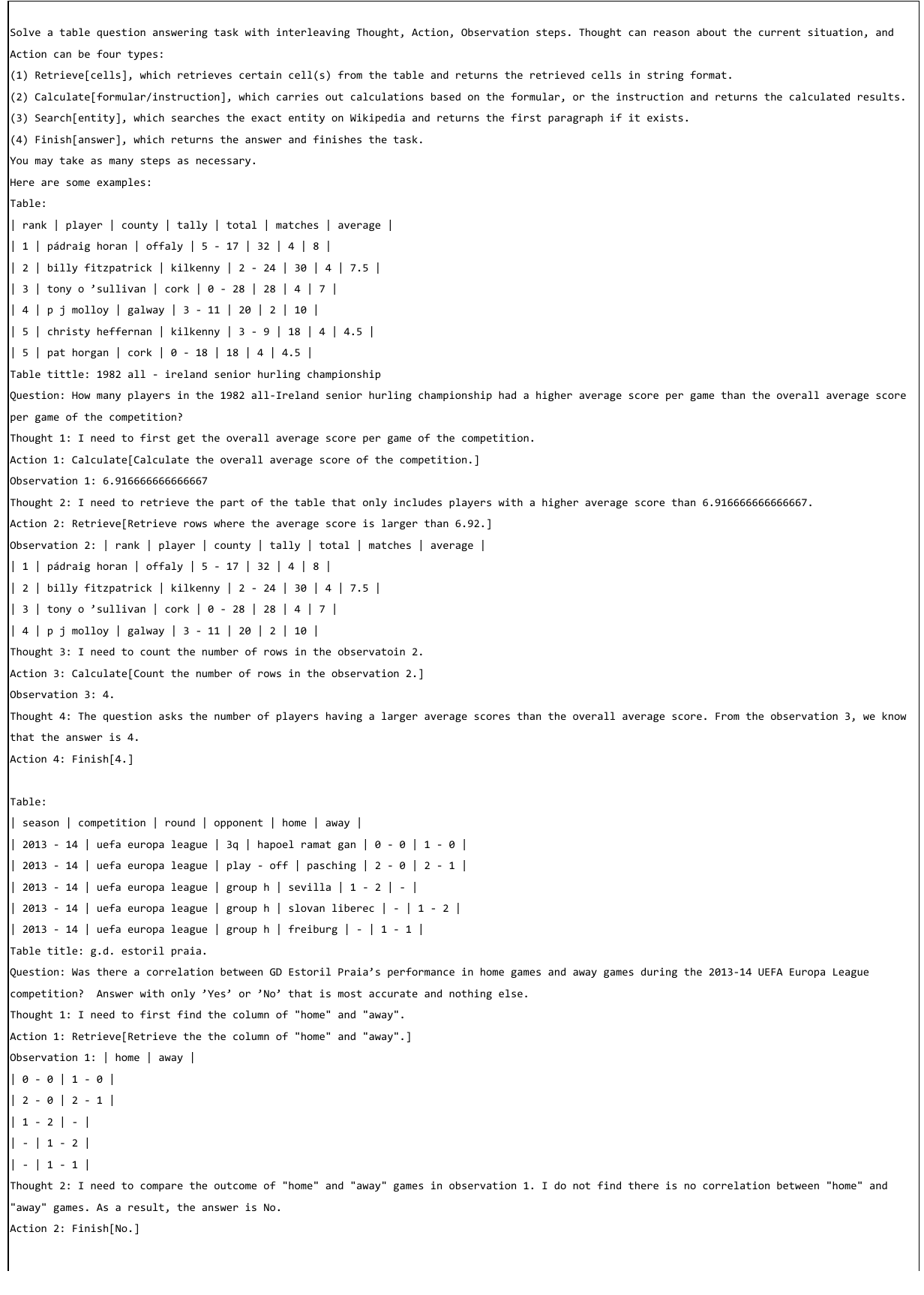}
\end{figure*}

\begin{figure*}[!ht]
    \centering
    \includegraphics[width=1.\textwidth]{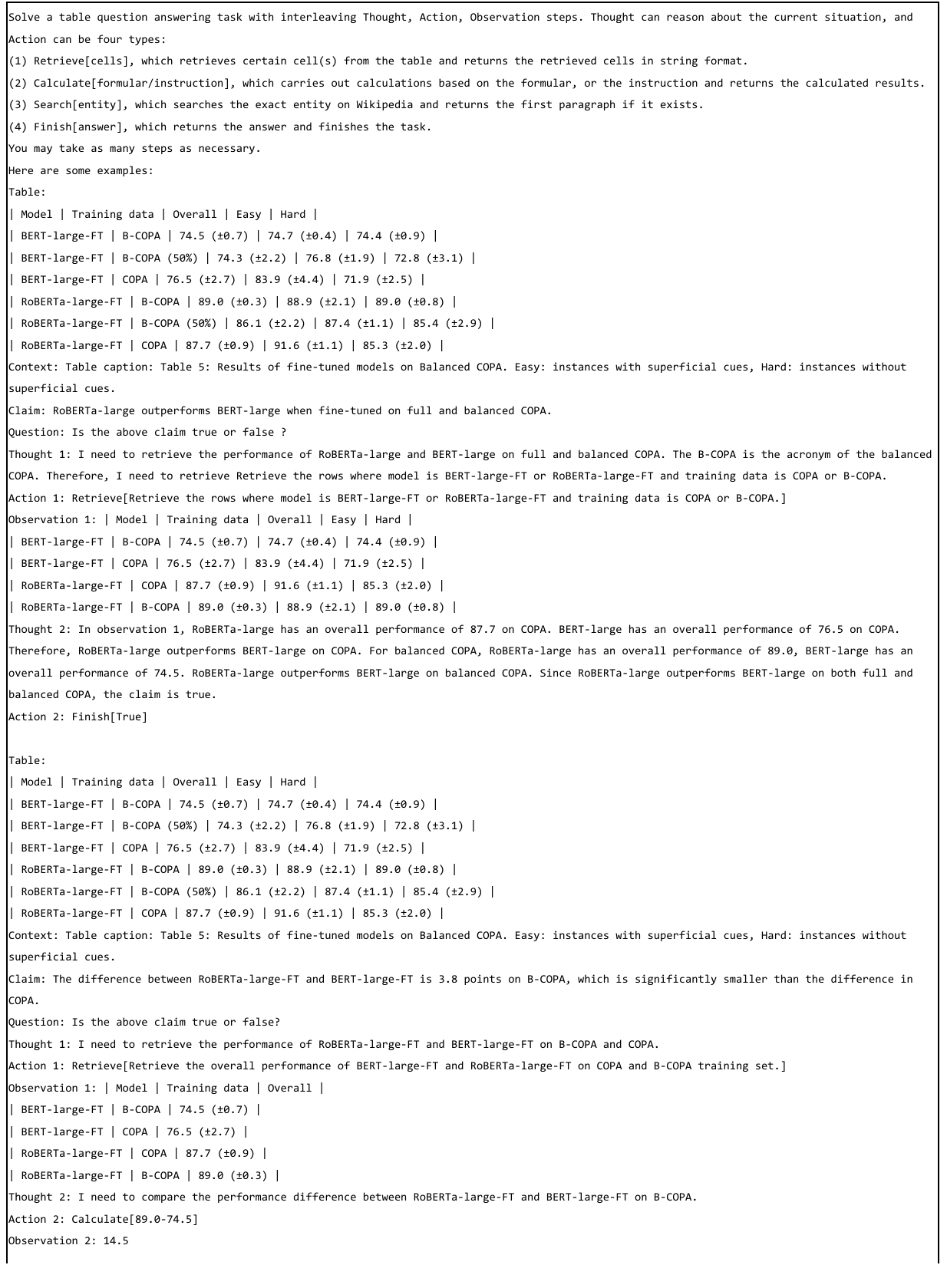}
\end{figure*}

\label{sec:appendix}

\end{document}